\documentclass[runningheads]{llncs}

\usepackage{graphicx}

\usepackage{lipsum}

\usepackage{amsmath} 
\usepackage{amssymb}
\usepackage{makecell}
\usepackage{multirow}
\usepackage{booktabs}
\usepackage{bbding}
\usepackage[marginal]{footmisc}

\usepackage{subfigure}
\usepackage{cite}
\usepackage{float}
\usepackage{color}

\begin{document}

\title{A Clinical-oriented Multi-level Contrastive Learning Method for Disease Diagnosis in Low-quality Medical Images}
\titlerunning{A Multi-level Contrastive Learning Method for Disease Diagnosis}

\author{Qingshan Hou\inst{1,2} \and Shuai Cheng\inst{1,2} \and Peng Cao\inst{1,2,3,(}\Envelope\inst{) }\and Jinzhu Yang\inst{1,2,3,(}\Envelope\inst{) }\and Xiaoli Liu \inst{1,2}\and Osmar R. Zaiane\inst{4} \and Yih Chung Tham\inst{5}}
\authorrunning{Q. HOU et al.}

\institute{Computer Science and Engineering, Northeastern University, Shenyang, China \and
Key Laboratory of Intelligent Computing in Medical Image of Ministry of Education, Northeastern University, Shenyang, China
\and National Frontiers Science Center for Industrial Intelligence and Systems Optimization, Shenyang 110819, China\\
\email{caopeng@mail.neu.edu.cn}\\
\email{yangjinzhu@cse.neu.edu.cn}\\
\and Alberta Machine Intelligence Institute, University of Alberta, Edmonton, Canada\\
\and  Ophthalmology, Yong Loo Lin School of Medicine, National University of Singapore, Singapore\\
}

\maketitle

\begin{abstract}
Representation learning offers a conduit to elucidate distinctive features within the latent space and interpret the deep models. However, the randomness of lesion distribution and the complexity of low-quality factors in medical images pose great challenges for models to extract key lesion features. Disease diagnosis methods guided by contrastive learning (CL) have shown significant advantages in lesion feature representation. Nevertheless, the effectiveness of CL is highly dependent on the quality of the positive and negative sample pairs. In this work, we propose a clinical-oriented multi-level CL framework that aims to enhance the model's capacity to extract lesion features and discriminate between lesion and low-quality factors, thereby enabling more accurate disease diagnosis from low-quality medical images. Specifically, we first
construct multi-level positive and negative pairs to enhance the model's comprehensive recognition capability of lesion features by integrating information from different levels and qualities of medical images. Moreover, to improve the quality of the learned lesion embeddings, we introduce a dynamic hard sample mining method based on self-paced learning. The proposed CL framework is validated on two public medical image datasets, EyeQ and Chest X-ray, demonstrating superior performance compared to other state-of-the-art disease diagnostic methods.

\end{abstract}

\section{Introduction}

Medical image classification plays a crucial role in clinical disease diagnosis. Automatically identifying whether medical images indicate health or disease, and even pinpointing specific illnesses, can alleviate the repetitive burden on clinicians and increase the efficiency of diagnoses. 
Though recent studies have shown that deep learning techniques hold promise for medical imaging applications~\cite{jiang2023review}, these techniques are often constrained by limited data annotations and insufficient supervision.

\begin{figure}[htbp]
    \centering
    \includegraphics[width=0.8\textwidth]{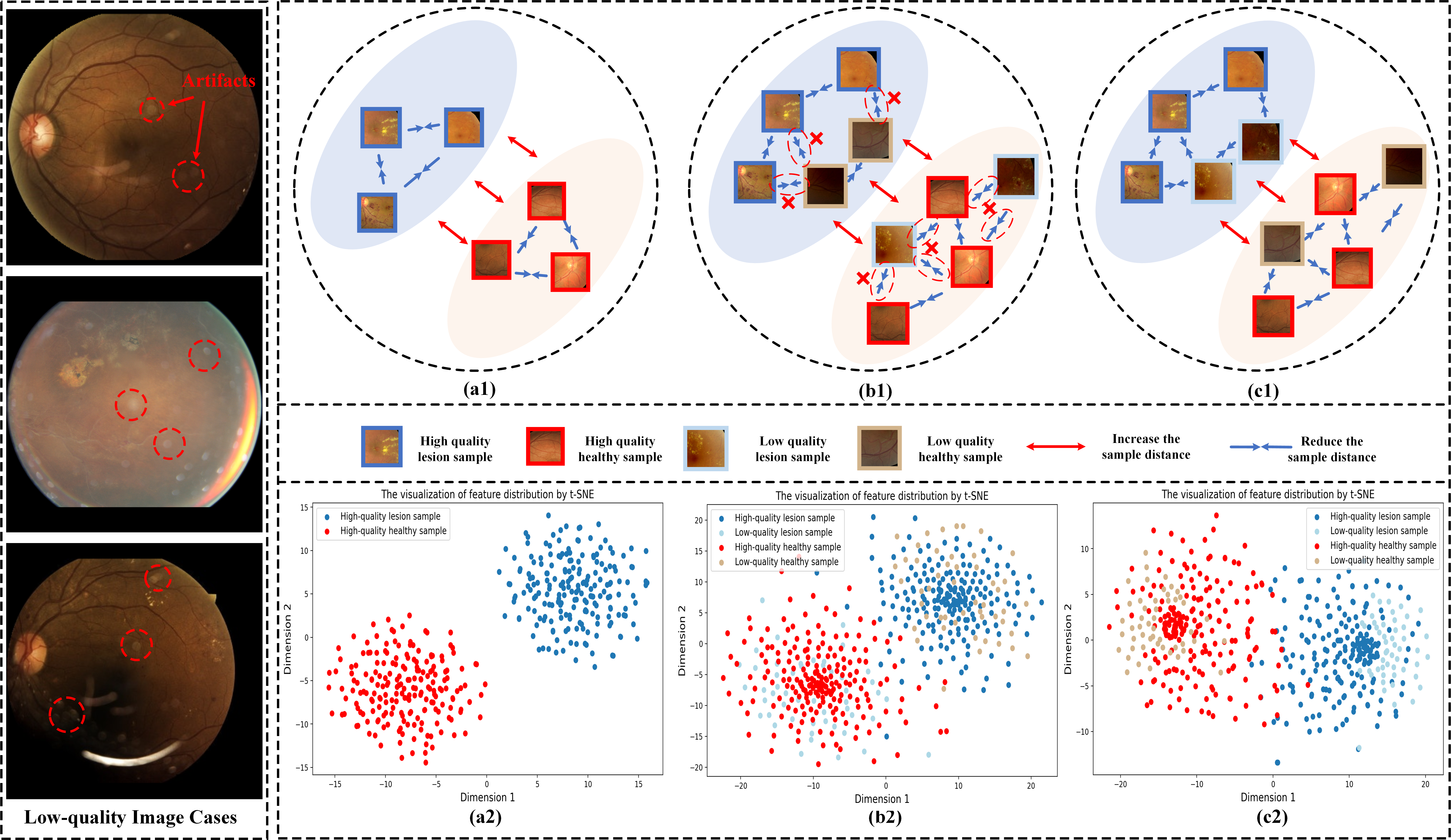}
    \caption{ \emph{Left}: Some cases of low-quality image. \emph{Right}:  
    We use t-SNE to visualize the feature distribution, with 200 high-quality lesion samples, 200 high-quality healthy samples, 50 low-quality lesion samples, and 50 low-quality healthy samples.
    (a1)\&(a2)-General contrastive learning with only high-quality images and the visualization of corresponding feature distribution. (b1)\&(b2)-The impact of low-quality factors on general contrastive learning and the visualization of feature distribution. (c1)\&(c2)-The proposed contrastive learning and the visualization of feature distribution.}
    \label{fig0}
\end{figure}

To address these challenges in real clinical settings and fully exploit medical images without pixel-level annotations, some existing studies~\cite{clindr1,clindr2,huang2021lesion} proactively explore the impact of contrastive learning (CL)~\cite{Moco,simCLR} on automated disease diagnosis models. 
However, they do not fully consider the common quality variations in medical images, which limits their effectiveness in eliminating the interference of low-quality factors on disease diagnosis. 
As shown in Figure~\ref{fig0}~\emph{left}, the medical images often suffer from various low-quality factors such as artifacts, blurring and so on, leading to a quality degradation~\cite{philip2005impact,anand2023chest}.
Ideally, CL guides diagnostic models to effectively distinguish between lesion samples and healthy samples in Figure~\ref{fig0}(a1)\&(a2). However, as illustrated in Figure~\ref{fig0}(b1)\&(b2), low-quality factors may cause CL to incorrectly pull the distance in the embedding space between lesion samples and low-quality healthy samples, or between healthy samples and low-quality lesion samples, thereby degrading the diagnostic performance of diseases.

The challenge mentioned above motivates us to develop a \textbf{C}linical-oriented \textbf{M}ulti-level \textbf{C}ontrastive \textbf{L}earning method, named CoMCL, tailored for automatic disease diagnosis on low-quality medical images. 
In the deployment of CL for disease diagnosis, our objective is to mitigate the effects of low-quality factors and false negative samples~\cite{alignment,crossCLR}, and to ensure that samples with similar semantic information remain close in the joint embedding space, as shown in Figure~\ref{fig0}(c1)\&(c2). 
To achieve this, we construct multi-level positive and negative pairs for the following three purposes: 1) Enhancing the ability of CL to distinguish low-quality factors from lesions in low-quality medical images; 2) Improving the ability of CL to discriminate between lesion and non-lesion areas; 3) Enhancing the awareness of CL to identify lesion characteristics in low-quality images. These abilities facilitate the modeling of the potential lesion-related embeddings, hereby enhancing the performance of the downstream diagnostic tasks.
In summary, our contributions can be summarized as follows. 
(1) We propose a multi-level CL framework that focuses on alleviating the impact of low-quality factors on lesion feature extraction. 
Besides, it also alleviates the issue of false negatives that occur when CL is introduced into the diagnosis of low-quality medical images.
(2) To improve the capability of CL in extracting lesion-related embeddings in low-quality medical images, we introduce a self-paced learning strategy to fully exploit and leverage hard negatives.
(3) CoMCL is evaluated on the large-scale {EyeQ} and Chest X-ray datasets. Experimental results show that CoMCL significantly outperforms the state-of-the-art automatic disease diagnosis methods. The code is available at ***.

\begin{figure}[htbp]
    \vspace{-0.5cm}
    \centering
    \includegraphics[width=0.9\textwidth]{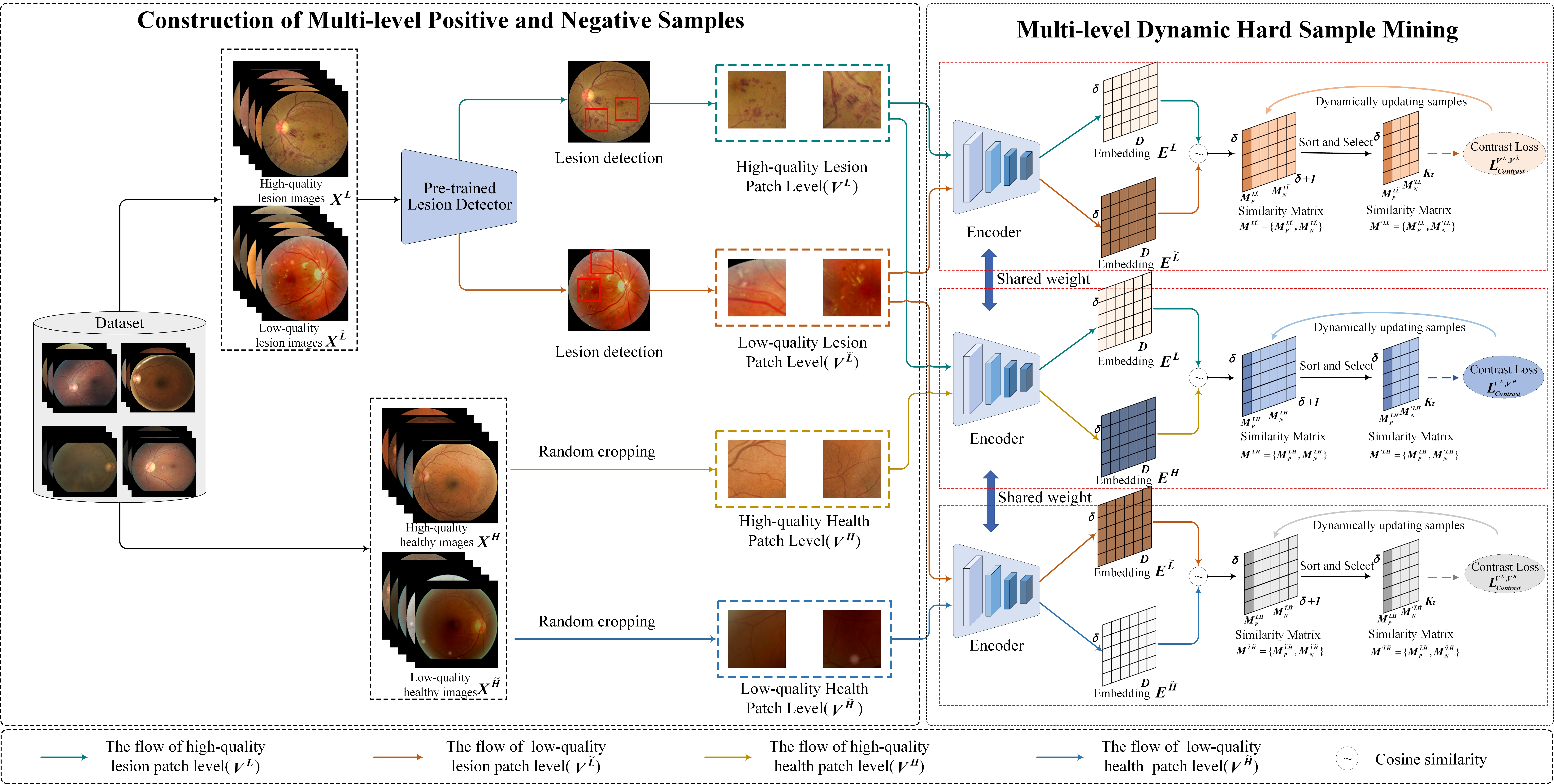}
    \caption{ The overall architecture of CoMCL comprises two components. 1)  Construction of multi-level positive and negative pairs for mitigating the effects of low-quality factors and false negatives. 2) Multi-level dynamic hard sample mining for improving the quality of the learned lesion-related embeddings.
    }
    \label{fig1}
    \vspace{-0.5cm}
\end{figure}

\section{Method}
Figure~\ref{fig1} shows the overview of CoMCL. In the first phase, we construct multi-level positive and negative pairs based on a lesion detector~\cite{huang2021lesion} pre-trained on an auxiliary dataset (IDRiD~\cite{IDRiD}) with pixel-level lesion annotations. This framework focuses on representing shared information among multi-level positive and negative pairs, alleviating the influence of imaging quality factors.
In the second phase, a self-paced learning-based dynamic sampling method effectively leverages hard negatives~\cite{hardsample,cai2020all} and enhances the learned feature embedding quality.

\subsection{Construction of multi-level positive and negative pairs}
In this section, we provide a comprehensive description of the multi-level construction of positive and negative pairs. The incorporation of the multi-level positive and negative pairs serves the dual purpose of aligning samples with similar semantic features and mitigating the effects of false negatives on the feature embeddings learned by the model. Specifically, given a dataset $X$ annotated with disease and quality annotations, we segregate $X$ into four subsets, including high-quality lesion images subset $X^{L}$, low-quality lesion images subset ${X}^{\widetilde{L}}$, high-quality healthy image subset $X^{H}$ and low-quality healthy image subset ${X}^{\widetilde{H}}$.
Then, we apply a pre-trained detector $f_{\text{det}}(\cdot )$ on $X^{L}$/${X}^{\widetilde{L}}$  and obtain high-confidence detection regions. Finally, the different level samples, denoted as $V^{m}=\{m\in \{L,\widetilde{L},H,\widetilde{H}\}|v_{1}^{m},v_{2}^{m} \dots v_{i}^{m}\}$, includes two parts: $\Omega(f_{\text{det}}(X^{m})>conf)$ ($m \in \{L,\widetilde{L}\}$) and $RandC(X^{m})$ ($m \in \{H,\widetilde{H}\}$), where \emph{conf} denotes the confidence threshold of detection results, $\Omega(\cdot)$ indicates the operation of expanding the predicted boxes of $f_{\text{det}}(\cdot)$ to 128*128 to ensure the inclusion of lesions as much as possible, and $RandC(\cdot)$ indicates randomly cropping images into patches with 128*128 from the healthy images.

Given a patch $v_{i}^{m}$ generated from $V^{m}$, we consider $\widetilde{v_{i}^{m}}$ that is an augmented version from $v_{i}^{m}$ as a positive sample and every patch $v_{k}^{n}$ in the $V^{n|n!=m}$ as negatives. Upon encoding each positive and negative sample, we acquire the feature embedding matrixes $e_i^{m}$/$\widetilde{e_i^{m}} \in E^m$ and $e_k^{n} \in E^n$, respectively. Subsequently, the contrastive loss can be defined as:
\begin{equation} \label{c1}
\begin{aligned}
L_{\text{Contrast}}^{V^m,V^n} & =-\sum_{i} \log \left(\frac{\exp \left(\operatorname{sim}\left(e_{i}^{m}, \widetilde{e_{i}^{m}}\right) / \tau\right)}{\exp(\operatorname{sim}(e^{m}_{i}, \widetilde{e_{i}^{m}})/ \tau )+\sum_{k} \exp \left(\operatorname{sim}\left(e^{m}_{i}, e_{k}^{n}\right) / \tau\right)}\right). 
\end{aligned}
\end{equation}

The primary goal of CoMCL is to enhance the model's ability to handle challenges arising from variations in image quality and concentrate on the accurate extraction of lesion-related features in the complicated clinical setting.
To this end, $V^{L}$ and its augmented versions are designated as positives, while $V^{\widetilde{L}}$ is identified as negatives, thereby deriving a CL loss, named $L_{\text{Contrast}}^{V^{L},V^{\widetilde{L}}}$. 
This constraint enables the model to better discriminate low-quality factors in medical images and minimize the influence of low-quality factors, leading to a more accurate embedding of lesion-related features.
Then, to further facilitate the extraction of lesion features and ensure the model's capability to differentiate between the lesion and healthy patches, we regard samples from $V^{L}$ as positive samples while $V^{H}$ as negatives, and define a CL loss, named $L_{\text{Contrast}}^{V^{L}, V^{H}}$.
Finally, to further improve the model's robustness and enhance its ability to distinguish  lesion from non-lesion regions in conditions of poor image quality, we treat samples from $V^{\widetilde{L}}$ as positives and samples from $V^{\widetilde{H}}$ as negatives, and devise a CL loss, named
$L_{\text{Contrast}}^{V^{\widetilde{L}},V^{\widetilde{H}}}$.

\subsection{Multi-level dynamic hard negatives mining}

In various supervised or unsupervised algorithms~\cite{kaya2019deep, birodkar2019semantic} based on metric learning, research on the impact of mining hard negatives on training suggests that not all negatives hold equal value in CL.
Moreover, hard negatives are semantically more similar to positives than regular negatives, indicating that hard negatives contain features of higher learning value, offering more potential beneficial information for CL.
Based on this finding, we introduce self-paced learning into CL. Given the model parameters $w$ at the current training step $t$, when updating the parameters of the model, we incorporate a binary variable $s_{i}$ based on the previous loss $L_{\text{Contrast}}^{V^{m},V^{n}}$, to decide whether each sample is selected. 
{According to the similarity matrix $M^{mn}=\{M_P^{mn}, M_N^{mn}\}$, the resampled sample matrix ${M}^{'mn}=\{M_P^{mn}, M_N^{'mn}\}$ can be defined as:

\begin{equation}
{M}^{'mn}=\left\{{z}_{i}^{m} \mid {z}_{i}^{m} \in Sort({M}^{mn}), \operatorname{sim}\left(z_{i}^{m}, {z}_{k}^{n}\right) \ge \operatorname{sim}(z_{i}^{m}, {z}_{K_{t} }^{n}\right) \}, 
\end{equation}
where $K_{t}$ is an adaptive parameter, determining the number of hard negatives to be considered. Specifically, $K_{t} = \lfloor \delta* cos(\frac{\pi t}{2  T_{max}} ) \rfloor $, where $\delta$ indicates the total number of negatives and $T_{max}$ is the maximum training step.
Therefore, for the update of multi-level CL models, we define the following optimization objective:}

\begin{equation} \label{c2}
\begin{aligned}
\left(w_{t+1}, v_{t+1}\right)=\operatorname{argmin}\left(r(w_{t})+\sum_{i=1}^{n} s_{i} L_{\text{Contrast}}^{V^m,V^n}({M}^{'mn})-\frac{1}{K_{t}} \sum_{i=1}^{n} s_{i}\right),
\end{aligned}
\end{equation}
{where $r(\cdot)$ denotes a regularization item, preventing the model from overfitting. By adjusting the value of $K_{t}$, we can adjust the number of hard negatives that affect the training procedure and model's generalization.}
If {$L_{\text{Contrast}}^{V^m,V^n}({M}^{'mn}) < \frac{1}{K_{t}}$}, then $s_{i}=1$ indicates that the sample is chosen for model fine-tuning. 
Otherwise, $s_{i}=0$ indicates that the sample is not selected. 
To validate the effectiveness of CoMCL, the parameters obtained from the multi-level CL phase are transferred to downstream disease diagnostic models, and then fine-tuned in a supervised learning setting to adapt to specific disease diagnostic tasks.

\section{Experiments}
\subsection{Datasets and Implementation Details}
\textbf{EyeQ dataset~\cite{EyeQ}} is a large public fundus image benchmark for diabetic retinopathy (DR) grading and quality assessment, containing 12,543 training and 16,249 testing images. Based on image quality and DR severity, the images are classified into three quality categories and five severity levels.\\
\textbf{Chest X-ray dataset~\cite{wang2017chestx}} is obtained from the public NIH-ChestXray14 multi-label dataset. The training and test sets contain 8,573 and 7,007 frontal X-ray images, respectively. Based on image quality and chest diseases, the dataset includes two quality labels: high and low, and eight disease types. We validate the promotive effect of CoMCL on multi-class disease diagnosis using this dataset.\\
\textbf{Implementation Details.} ResNet50~\cite{resnet} is used as the backbone network for feature extraction, with the global average pooling and fully connected layers removed. For the construction of multi-level positive and negative pairs, all patches are cropped to 128×128 due to varying original image sizes. 
The temperature parameter $\tau$ in Equation~\ref{c1} is set to $0.07$. 
During multi-level dynamic hard sample mining, parameters are optimized using the Adam optimizer (momentum=0.9) over 800 epochs. Training starts with a learning rate of $1\times10^{-3}$ and a batch size of 400. 
\subsection{Comparison with the State-of-the-Art}
This section presents both quantitative and qualitative comparisons with various recent disease diagnostic methods on the EyeQ and Chest X-ray datasets, showcasing the effectiveness of the CoMCL framework for single-label (DR grading) as well as multi-label chest disease diagnosis.
We compare CoMCL with several comparable methods, including Resnet50, Inception-v3, DenseNet-121, MMCNN~\cite{MMCNN}, Zoom-in-Net~\cite{zoominnet}, Lesion-base CL~\cite{huang2021lesion}, CABNet~\cite{9195035}, DeepMT-DR~\cite{DeepMT}, Lesion-aware CL~\cite{Lesion-aware}, and LANet~\cite{hou2023diabetic}. 
For all comparable methods, we follow the same experimental setup described in their original papers to ensure the fairness and competitiveness of each competing approach. 

\begin{table}[htbp]
  \centering
  \vspace{-0.2cm}
  \renewcommand{\arraystretch}{1}
  \caption{The comparison between CoMCL and the comparable methods in DR grading and multi-label chest disease diagnosis.}
  \resizebox{\linewidth}{!}{ 
    \begin{tabular}{l|cc|cc|cc|cc|cc|cc}
    \toprule
    \multicolumn{1}{c|}{\multirow{3}[0]{*}{Methods}} & \multicolumn{6}{c|}{EyeQ Dataset}             & \multicolumn{6}{c}{Chest X-ray Dataset} \\
\cline{2-13}          & \multicolumn{2}{c|}{33.4\%} & \multicolumn{2}{c|}{70\%} & \multicolumn{2}{c|}{100\%} & \multicolumn{2}{c|}{43.2\%} & \multicolumn{2}{c|}{70\%} & \multicolumn{2}{c}{100\%} \\
\cline{2-13}          & \multicolumn{1}{c}{Kappa} & \multicolumn{1}{c|}{ACC} & \multicolumn{1}{c}{Kappa} & \multicolumn{1}{c|}{ACC} & \multicolumn{1}{c}{Kappa} & \multicolumn{1}{c|}{ACC} & \multicolumn{1}{c}{Kappa} & \multicolumn{1}{c|}{ACC} & \multicolumn{1}{c}{Kappa} & \multicolumn{1}{c|}{ACC} & \multicolumn{1}{c}{Kappa} & \multicolumn{1}{c}{ACC} \\
    \hline
    Resnet50 &   0.804    &   0.783    &   0.743    &   0.715    &   0.674    &    0.662   &   0.619    &   0.626    &   0.594    &   0.613    &   0.564    & 0.589 \\
    Inception-v3 &   0.798    &  0.776     &   0.728    &   0.706    &   0.652    &    0.637   &  0.615     &  0.624     &    0.587   &  0.609     &    0.558   & 0.576 \\
    DenseNet-121 &   0.813    &   0.794    &   0.756    &   0.732    &   0.683    &   0.668    &   0.624    &   0.635    &    0.603   &   0.658    &   0.572    & 0.607 \\
    MMCNN~\cite{MMCNN} &    0.862   &   0.841    &    0.795   &   0.778    &  0.725     &  0.704     &   0.657    &   0.672    &   0.632    &   0.644    &   0.604    & 0.626 \\
    Zoom-in-Net~\cite{zoominnet} &   0.873    &   0.854    &  0.812     &   0.784    &  0.736     &    0.713   &   0.662    &   0.684    &   0.648    &  0.653     &   0.615    & 0.631 \\
    Lesion-base CL~\cite{huang2021lesion} &   0.848    &   0.832    &    0.783   &   0.761    &   0.694    &    0.672   &   0.636    &   0.652    &   0.617    &   0.628    &   0.584    & 0.619 \\
    CABNet~\cite{9195035} &  0.865  & 0.847 & 0.797 & 0.782  &  0.731  & 0.709 & 0.660 & 0.675  &  0.643  & 0.651 & 0.607 & 0.628 \\
    DeepMT-DR~\cite{DeepMT} &    0.857   &    0.839   &   0.791    &   0.768    &   0.712    &  0.694     &   0.649    &   0.665    &    0.621   &   0.639    &   0.592    & 0.624 \\
    Lesion-aware CL~\cite{Lesion-aware} &   0.876    &   0.857    &   0.824    &   0.816    &   0.757    &   0.724    &  0.673     &    0.691   &  0.653     &   0.664    &   0.626    &  0.645\\
    LANet~\cite{hou2023diabetic} &  0.854  & 0.835 & 0.786  & 0.765  & 0.706   & 0.683   &  0.642  &  0.658  &  0.619  & 0.634  & 0.587  & 0.622 \\
    \hline
    CoMCL(Ours) &  \textbf{0.884}     &   \textbf{0.872}    &   \textbf{0.852}    &   \textbf{0.837}    &   \textbf{0.793}    &   \textbf{0.776}    &   \textbf{0.682}    &   \textbf{0.708}    &   \textbf{0.665}    &   \textbf{0.672}    &    \textbf{0.641}   &  \textbf{0.663} \\
    \bottomrule
    \end{tabular}%
    }
  \label{tab1}%
  \vspace{-0.3cm}
\end{table}%

As shown in Table~\ref{tab1}, CoMCL significantly outperforms comparable methods for both DR grading and multi-label chest disease diagnosis, achieving higher Kappa and Accuracy scores. The results highlight several interesting observations:  
(1) We first explore the impact of the proportion of low-quality images in the EyeQ and Chest X-ray datasets on CoMCL and comparable methods. As the proportion of low-quality images increases from the original 33.4\%/43.2\% (the proportion of low-quality images in the original dataset) to 100\% (by degrading all images using ~\cite{annotation1, li2022structure}), the performance of all diagnostic methods decreases across datasets, indicating the negative impact of low-quality factors. However, CoMCL exhibits a stronger ability to avoid interference from low-quality factors compared to other methods.
(2) Compared to previous CL methods (Lesion-base CL and Lesion-aware CL), CoMCL exhibits significant advantages across datasets under different proportions of low-quality images. This advantage mainly benefits from the fact that CoMCL incorporates information from multiple levels and different qualities of medical images. Therefore, CoMCL acquires lesion features that are not obvious in low-quality patches, mitigating the influence of low-quality factors on the learned lesion embeddings, and enhancing the model's ability to comprehensively identify lesion-related features.

\subsection{Ablation Study}
 
To comprehensively investigate the contribution of multi-level positive and negative pairs  and self-paced learning to disease diagnosis, we compare CoMCL with several variants:
(1) Baseline (Resnet 50): Training a basic classification model on the EyeQ dataset.
(2) Basic CL: Pre-training a basic CL model~\cite{Moco} and fine-tuning the downstream classification model on EyeQ.
(3) CoMCL w/o Multi-level: Constructing positives and negatives using only lesion and healthy samples respectively, without the construction of multi-level positive and negative pairs.
(4) CoMCL w/o SPL: Training without considering self-paced learning, i.e., without considering hard negatives.

\begin{table}[h]
\vspace{-0.2cm}
\centering
\begin{minipage}[h]{0.48\textwidth}
\centering
\renewcommand{\arraystretch}{1}
\caption{Ablation study of CoMCL on EyeQ dataset.}
\begin{tabular}{l l l}
\hline
Methods & ACC & Kappa \\
\hline
Baseline(Resnet 50) & 0.804  & 0.783  \\
Basic CL & 0.842  & 0.827  \\
CoMCL w/o Multi-level & 0.859  & 0.836  \\
CoMCL w/o SPL & 0.868  &  0.860 \\
CoMCL & \textbf{0.884} & \textbf{0.872}   \\
\hline
\end{tabular}
\label{tab2}
\end{minipage}
\hfill
\begin{minipage}[h]{0.48\textwidth}
\centering
\renewcommand{\arraystretch}{1}
\caption{Performance of CoMCL on EyeQ under different combinations.}
    \begin{tabular}{l|cccc|cc}
    \hline
    Method & \multicolumn{1}{l}{{\textit{$V^{L}$}}} & \multicolumn{1}{l}{{\textit{$V^{\tilde{L}}$}}} & \multicolumn{1}{l}{{\textit{$V^{H}$}}} & \multicolumn{1}{l|}{{\textit{$V^{\tilde{H}}$}}} & \multicolumn{1}{l}{ACC} & \multicolumn{1}{l}{Kappa} \\
    \hline
    {CoMCL\_v1}& \checkmark     &   \checkmark    &       &       &  0.846     &  0.835\\
    {CoMCL\_v2}&   \checkmark    &   \checkmark    &   \checkmark    &       &   0.867   & 0.856 \\
    {CoMCL\_v3}&   \checkmark    &    \checkmark   &   \checkmark    &   \checkmark    &    \textbf{0.884}   &  \textbf{0.872} \\
    \hline
    \end{tabular}%
  \label{tab4}%
\end{minipage}
\vspace{-0.3cm}
\end{table}

The experimental results are shown in Table~\ref{tab2}. The following aspects can be revealed: 
1) The baseline model exhibits the lowest performance, underscoring the importance of CL in modeling specific lesion embeddings. Furthermore, compared to other variants of CoMCL, the baseline CL performs the worst, reflecting the adverse impact of low-quality factors on contrastive learning and the significance of hard negatives to contrastive learning.
2) The performance of CoMCL w/o Multi-level is lower than that of CoMCL. The construction of multi-level positive and negative pairs is crucial for improving diagnostic performance when introducing CL for automatic disease diagnosis. By considering the impact of low-quality factors on lesion-related embedding extraction, CoMCL can learn relevant lesion embeddings more effectively, thereby achieving improved disease diagnostic performance. 
3) The performance of CoMCL w/o SPL is lower than CoMCL.  
By dynamically mining hard negatives through self-paced learning, the quality of lesion embeddings is further improved, thereby enhancing the performance of disease diagnostic tasks.

{Subsequently, we investigate the impact of incorporating different levels on the final result, as depicted in Table~\ref{tab4}, which shows the performance comparison of the CoMCL method under varying level combinations. 
From the results, it is clear that the model CoMCL\_v3, by considering all level patches, achieves the best results in terms of accuracy and kappa. 
The experimental results show that simultaneously considering the learning of discriminative embeddings for lesions and the identification of quality factors achieves a more significant performance boost after fine-tuning the downstream tasks.
Additionally, incorporating low-quality levels can enhance the model's ability to discern lesion features under limited imaging conditions, ensuring that the model's diagnostics do not solely rely on high-quality image features.
This approach ensures the effectiveness of the algorithm under complex clinical imaging conditions.}

\begin{figure}[htbp]
    \vspace{-0.3cm}
    \centering
    \includegraphics[width=0.9\textwidth]{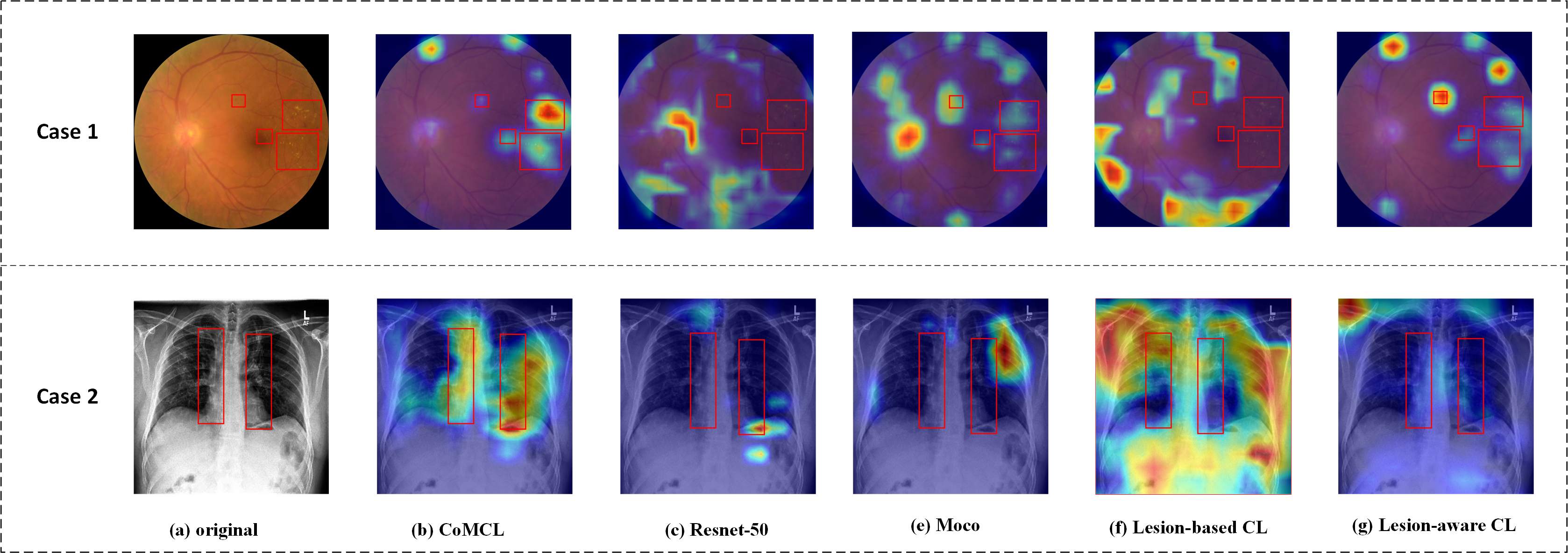}
    \caption{ Visualization results of Regions of Interest (RoIs) across the ResNet and the representative CL methods.}
    \label{fig3}
    \vspace{-0.5cm}
\end{figure}
{Figure~\ref{fig3} illustrates the visualization results of baseline(Resnet50) and different contrastive learning-based diagnostic methods (Moco, Lesion-base CL and Lesion-aware CL) for two medical cases: a fundus image indicative of proliferative diabetic retinopathy (Case 1) and a chest X-ray showing potential pulmonary pathologies (Case 2). 
While the other methods are capable of capturing lesions to a certain extent, they are distracted by more prominent physiological features and high-contrast edges. In contrast, CoMCL more clearly emphasizes lesion regions.
This indicates that CoMCL has the ability to distinguish lesion-relevant features from structural aspects and complex low-quality factors of the medical image, which is crucial for accurate disease diagnosis.}

\section{Conclusion}
In this study, we propose a clinical-oriented multi-level contrastive learning framework for disease diagnosis in low-quality medical images. The proposed framework, by constructing multi-level positive and negative pairs, can explore lesion features from different levels, thereby mitigating the influence of low-quality factors on the model's extraction of lesion features. Additionally, we design a dynamic hard negative mining scheme based on self-paced learning to fully utilize hard negative samples, significantly improving the quality of feature embedding. Experimental results show that CoMCL significantly improves the accuracy of disease diagnosis in low-quality medical images, which is crucial for conserving medical resources and enhancing the efficiency of medical services.

\clearpage
\bibliographystyle{splncs04}
\bibliography{refs}
\end{document}